\documentclass[11pt,letterpaper]{article}
\usepackage{cogsys}
\usepackage[T1]{fontenc}
\usepackage{times}
\usepackage[pdftex]{graphicx} 

\usepackage{graphicx}
\usepackage{subcaption}
\usepackage[dvipsnames]{xcolor}

\usepackage{natbib}
\usepackage{amsmath}
\usepackage{amsfonts}
\usepackage[ruled,vlined]{algorithm2e}
\usepackage{algpseudocode}
\usepackage{booktabs}
\usepackage{multirow}
\usepackage{amsmath,amssymb}
\usepackage{amsthm}
\newtheorem{definition}{Definition}
\newcommand{\smallpm}[1]{\mathbin{\scriptstyle\pm\scriptstyle#1}}

\cogsysheading{X}{2025}{1-6}{X/20XX}{X/20XX}

\ShortHeadings{Human Allied RRL}
              {F. Golivand et al.}

\begin{document}

\title{Human-Allied Relational Reinforcement Learning}

\def\thefootnote{*}\footnotetext{These authors contributed equally.}\def\thefootnote{\arabic{footnote}}
 
\author{Fateme Golivand\footnotemark[1]}{Fateme.GolivandDarvishvand@utdallas.edu}
\author{Hikaru Shindo$^*$\footnotemark[2]}{Hikaru.Shindo@cs.tu-darmstadt.de}
\author{Sahil Sidheekh$^*$\footnotemark[1]}{Sahil.Sidheekh@utdallas.edu}
\author{Kristian Kersting\footnotemark[2]}{Kristian.Kersting@cs.tu-darmstadt.de}
\author{Sriraam Natarajan\footnotemark[1]}{Sriraam.Natarajan@utdallas.edu}
\address{\footnotemark[1] The University of Texas at Dallas, Richardson, TX, USA\\
\footnotemark[2] Technical University Darmstadt}
\begin{abstract}
Reinforcement learning (RL) has experienced a second wind in the past decade. While incredibly successful in images and videos, these systems still operate within the realm of propositional tasks ignoring the inherent structure that exists in the problem. Consequently, relational extensions (RRL) have been developed for such structured problems that allow for effective generalization to arbitrary number of objects. However, they inherently make strong assumptions about the problem structure. We introduce a novel framework that combines RRL with object-centric representation to handle both structured and unstructured data. We enhance learning by allowing the system to actively query the human expert for guidance by explicitly modeling the uncertainty over the policy. Our empirical evaluation demonstrates the effectiveness and efficiency of our proposed approach.
\end{abstract}

%

\section{Introduction}

Reinforcement learning (RL) agents learn to make decisions by interacting with their environment~\citep{sutton1998reinforcement}. Their ability to learn complex behavior without labeled training data has attracted significant attention, which has been magnified by the success of deep learning-based approaches to RL. These include game playing~\citep{mnih2013playing, silver2017masteringchessshogiselfplay, vinyals2017starcraftiinewchallenge}, enhancing Large Lanugage Models (LLMs)~\citep{wang2025reinforcementlearningenhancedllms}, robotics~\citep{kober2013reinforcement} and autonomous driving~\citep{10682977}.
While incredibly successful on specific tasks, deep learning based RL (DRL) suffers from a few limitations -- the learned models are not easy to interpret, they assume a flat, propositional feature space, and most importantly do not generalize well to an arbitrary number of objects and unseen tasks.

These limitations are addressed by the relational reinforcement learning (RRL) framework, which uses symbolic logic to represent states and actions.~\citep{tadepalli2002rrl,das2020fitted, kersting2008non, dvzeroski1998relational}. 
This rich representation not only allows agents to generalize to an arbitrary number of unseen objects and situations, but it also enhances their interpretability.
These models can explore at multiple levels of abstraction, i.e., at the level of objects (say {\em block}) or at the level of relations (for ex., {\em on(X,Y)}), at the level of sub-populations (for ex., {\em squareBlock}) or at the individual instance level (for ex., {\em b1}). Since they are built on symbolic manipulation methods from class AI literature, the powerful exploration results in effective learning and generalization.  The resulting policies address the twin challenges of interpretability by employing a rich symbolic representation, and generalization by employing universal and existential quantifiers ~\citep{natarajan2011imitation, das2020fitted, van2012solving,kersting2004bellman,  kersting2008non}.

While extremely powerful from a representation perspective, the applications of these methods were limited to structured domains. Specifically, they assume that the background knowledge, specific observations, language bias, and search bias are all provided before learning and are available in symbolic form. This has greatly limited the applicability of these systems to compelling tasks where the inputs are not necessarily structured. It is indeed challenging to learn from fully structured inputs -- for instance, blocksworld has not yet been solved in its full generality. However, constructing these structured information from fully unstructured data such as sequences of images or videos can be time-consuming and cumbersome. However, faithful modeling of such problems requires structured representation and learning from unstructured data.


We aim to address this problem of learning RRL agents without assuming structured domains and fully pre-specified domain knowledge. Specifically, we aim to answer the following question: {\em can we learn symbolic, policies in both structured and unstructured domains in the presence of a human with a fixed budget?} 
To answer this question, we propose the {\bf RAEL} (Relational Active Advice Elicitation) algorithm, which actively elicits advice from the human to learn symbolic policies for acting in both structured and unstructured environments. At a high-level, RAEL works as follows: It uses a symbolic representation as input and for tasks that do not have structured information (for instance, Atari games), uses an object-centric representation extractor method to create these symbolic representations \citep{nudge23}. Given this representation, the system employs Relational Fitted-Q learning (RFQ) \citep{das2020fitted} to learn the policy. While learning the policy, the algorithm explicitly computes the uncertainty over the policy based on policy roll-out and identifies the most uncertain states to query the human expert. {To control the frequency of such interactions and limit the burden on the expert, RAEL operates under a predefined advice budget, which constrains the total number of queries that can be made during training.} Given the advice from the human, the system trades-off between its learned policy and the human knowledge to identify the best action in every state. Essentially, this advice is integrated as a soft constraint to the learning process.

We make the following key contributions: (1) As far as we are aware, we present the first work on {\bf actively querying the human expert} for learning symbolic policies from both structured and unstructured data. (2) We go beyond shaping functions and directly obtain policy constraints from the user. (3) Our RAEL algorithm can learn from both unstructured and structured data in the presence of a human expert. (4) Our empirical evaluations on both structured and unstructured domains demonstrate the efficacy and efficiency of the proposed approach.

The rest of the paper is organized as follows -- after introducing the necessary background on RL, RRL and advice taking, we present our Relational Active Advice Elicitation (RAEL) framework. We then provide empirical evidence of the algorithm by comparing against strong baselines. Next, we present the related work before concluding the paper by outlining areas of future research.

\section{Background and Preliminaries}
\paragraph{Reinforcement Learning (RL)} 
\citep{sutton1998reinforcement} provides a principled framework for sequential decision making under uncertainty by modeling it as a \emph{Markov Decision Process (MDP)}. An MDP is formally defined as the $5-$tuple:  $\mathcal{M}=<\mathcal{S},\mathcal{A},P,R,\gamma>$, where $\mathcal{S}$ denotes the set of all possible states of an environment, $\mathcal{A}$ denotes the set of actions an agent can perform in that environment, $P: \mathcal{S} \times \mathcal{A} \times \mathcal{S} \to [0,1]$ denotes the transition probability function,
$R: \mathcal{S} \times \mathcal{A} \to \mathbb{R}$ denotes the reward function and $\gamma \in [0,1)$ is a discount factor that trades off immediate versus future rewards. At each time step $t$, the agent observes a state $s_t$, selects an action $a_t$, transitions into a new state $s_{t+1}$ according to $P(s_{t+1}|s_t,a_t)$ and receives a reward $r_t=R(s_t, a_t)$. The goal is to learn a policy $\pi(a|s)$, which specifies a probability distribution over the actions given a state. The performance of the policy can be summarized by an \textbf{action-value function} (a.k.a Q-function)
\(
Q^\pi(s,a) = \mathbb{E}_\pi \left[ \sum_{t=0}^{\infty} \gamma^t R(s_t, a_t) \big| s_0=s, a_0=a \right]
\),
which measures the expected cumulative discounted reward achieved by starting from state $s$, taking action $a$ and thereafter following the policy $\pi$. The optimal Q-function satisfies the Bellman optimality equation:
\(
Q^*(s,a) = \mathbb{E}_{s' \sim P(\cdot |s,a)} \left[ R(s, a) +\gamma \max_{a'} Q^*(s',a') \right]
\).
While dynamic programming can be used to solve this exactly in small and finite spaces,  
it becomes intractable when $\mathcal{S}$ or $\mathcal{A}$ is large or continuous. \textbf{Fitted Q-learning} \citep{ernst2005tree} addresses this by approximating the Q-function ($\widehat{Q}$) using a class of regression models. Specifically, it collects a batch of transitions $\mathcal{D}=\{ (s_i,a_i,r_i,s_i')_{i=1}^N\}$ under an exploration strategy, and at each iteration, computes the empirical Bellman targets
\(y_i = r_i + \gamma \max_{a'} \widehat{Q}(s_i',a')\).
A function class $\mathcal{F}$ (such as regression trees, neural networks, etc) is then fit to approximate the Q-function by minimizing empirical error as follows: 
\(
\widehat{Q} \in \arg \min_{f \in \mathcal{F}} \sum_i(f(s_i,a_i)-y_i)^2 + \Omega(f)
\).
where $f(s_i, a_i)$ denotes a parameterized approximation of the Q-function projected to the function class $\mathcal{F}$ and $\Omega$ denotes a regularizer that penalizes complexity. Thus, by iterating between the Bellman operator and function projection, Fitted Q-learning reduces the problem of approximating the optimal action-value function to an efficient supervised regression problem, and admits generalization and convergence guarantees under standard assumptions \citep{ernst2005tree}. 

\paragraph{Relational RL (RRL).}
Many real-world problems are most naturally represented in terms of objects and relationships between them, rather than flat feature vectors. While deep RL methods such as DQN \citep{mnih2015human} or PPO \citep{schulman2017proximal} provide strong function approximation in propositional settings, they are not designed to operate over relational state spaces without significant feature engineering (e.g., relational graph embeddings). Moreover, their gradient-based optimization often results in sample inefficiency, requiring large amounts of data to reach competitive performance, especially in structured/relational domains. RRL solves this issue by extending RL to relational domains. It models these domains using \textbf{Relational Markov Decision Process (RMDP)}~\citep{fern2003approximate}, which extends the classic MDP framework by representing states, actions, and transitions using first-order logic. 
Formally, the environment is characterized by a set of objects classes $(\mathcal{O})$, a set of predicate symbols $(\mathcal{P})$ (for relations and attributes), and a set of action types ($\mathcal{A}$). A state $s$ is specified by the set of ground, instantiated predicates (atoms) that are true for the current objects, for example, $\texttt{on(block1,block2)}$. The set of available ground actions in a state is obtained by instantiating the action types for the current objects. The transition probabilities and reward function are defined as in the standard MDP, but now operate over relationally described states and actions. A key challenge in RMDPs is that the number of ground states and actions grows combinatorially with the number of objects, and hence learning policies and value functions in tabular form is infeasible. RRL algorithms address this by learning a relational policy ($\pi(s,a)$), which is a mapping that assigns a probability to each ground action $a$ in state $s$, but is specified in a compact \textbf{lifted} (parameterized) form using logical rules over object types and relations, rather than enumerating all possible cases. RFQ~\citep{das2020fitted}, for example, computes the Bellman targets over batches of transitions described in terms of symbolic states and actions, and uses relational rule learners such as relational regression trees or their gradient-boosted ensemble versions to project the Q-values onto the space of relational functions. We will focus on RFQ with gradient-boosted relational regression trees as the function class ($\mathcal{F}$) in this work, as it offers a stable, interpretable and sample-efficient way to accurately approximate the Q-function, while allowing easy integration of human feedback.

\paragraph{Object Centric Representations.}
Typical RL policy functions are implemented using deep neural networks—such as convolutional neural networks—that take raw visual inputs (pixels) and output action distributions. However, studies have shown that these pixel-based policy functions are often \emph{misaligned}: they may develop biases and base decisions on misleading features, such as focusing on background elements rather than relevant foreground objects, thereby contradicting human intuitions about the environment~\citep{delfosse2024interpretable_concept_bottlenecks, delfosse2025deep}.
To address this limitation, \emph{object-centric representations} have emerged as a key component in interpretable RL~\citep{yoon2023ocr_rl,haramati2024entity_centric_rl}. These methods represent the input space in terms of objects (e.g., positions, orientations) and their relations (e.g., left of, right of), rather than undifferentiated pixels.
Thus, in this work, we employ them as a way to learn transparent policies that can utilize both symbolic and sub-symbolic information within relational gradient-boosted fitted Q-learning.

\paragraph{Human Advice in RL.}
In complex real-world environments, state and action spaces are often high-dimensional, and rewards from the environment can be sparse. Under such conditions, RL algorithms may suffer from slow convergence and poor sample efficiency, since discovering rewarding behaviors often requires exploring a large number of unproductive or even unsafe regions of the state space~\citep{arnold2021challenges}.
Human guidance has emerged as a compelling solution to address this problem, as providing expert advice can ease the optimization problem by reducing the search space, accelerating convergence, and steering agents away from undesirable behaviours \citep{Retzlaff2024HumanintheLoop}. While several works on reward-shaping and advice-taking \citep{ng1999policy,ng2000algorithms,kunapuli2013guiding,maclin1996} have shown that incorporating additional feedback can improve sample efficiency, they typically require heuristics or dense up-front input/advice from the expert, which can be expensive and infeasible in practice. 
To address this, \textbf{active advice seeking} \citep{odom2015active} has been proposed, where agents automatically identify situations of high uncertainty such as unfamiliar or ambiguous states, based on measures such as policy entropy, and selectively request targeted guidance from a human expert. By querying for advice only when needed, such systems can improve the policy quality using far fewer interactions with the environment, all the while staying within the budget at hand, making human-in-the-loop RL practical and scalable to large real world applications.

In this work, we build on the above foundations to introduce a unified framework that combines relational representations, fitted-Q learning, object-centric representations and active advice seeking to enable efficient, interpretable, and human-aligned RL in complex domains.

\begin{figure}[t]
\centering
\includegraphics[width=1.\columnwidth, keepaspectratio]{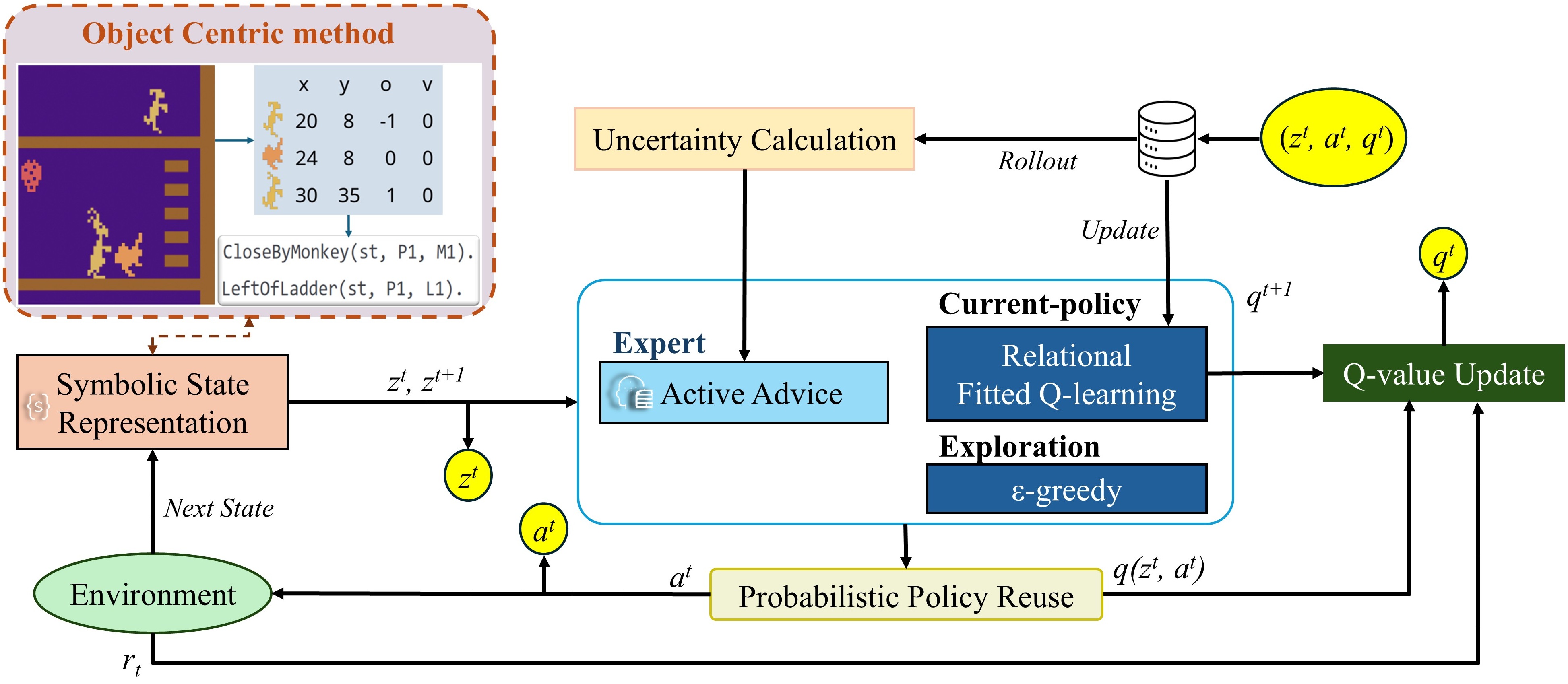} 

\caption{Overview of the proposed \textbf{Relational Active Advice Elicitation (RAEL)} framework. The environment provides either raw or symbolic states; an object-centric extraction module lifts raw images to symbolic representations. The agent’s behavior is determined by a probabilistic policy reuse strategy, combining relational fitted Q-learning with $\epsilon$-greedy exploration and active advice from an expert. After each training cycle, policy rollouts are used to compute statewise uncertainty. The most uncertain states are presented to the expert for advice, which is stored as abstractions for reuse. Advice-aware Q-value updates incorporate this guidance to improve learning efficiency.}
\label{fig:flowchart}
\end{figure}

\section{Relational Active Advice Elicitation in Reinforcement Learning (RAEL)}
The goal of Relational Active advice Elicitation in reinforcement Learning (RAEL) is to learn a policy that automatically detects problematic regions of the state space, actively solicits expert guidance on those areas, and—by accumulating advice—achieves greater sample efficiency, stability, and confidence in its decisions.
With RAEL, we aim to solve the following problem:
\begin{center}
\fbox{%
  \begin{minipage}{\dimexpr\linewidth-2\fboxsep-2\fboxrule\relax}
  \raggedright
  \textbf{Given:} An environment that yields either raw observations (e.g., images) or symbolic state descriptions, a set of (parameterized) action types $\mathcal{A}$, and a finite budget $B$ of advice from an expert.\\
    \textbf{Task:} Learn a policy $\pi$ that maximizes the expected return while (a) generalizing across varying number and configurations of objects, (b) is sample-efficient and stable, and (c) uses human advice only when needed.
  \end{minipage}%
}
\end{center}

To this end, we train an agent to act in a relational (first-order) state space, which represents each state primarily by objects' attributes and relations. If the environment provides raw observations (e.g., pixels), we map sub-symbolic inputs to symbolic states via
\(
f:\; x \mapsto z,\; x\in\mathbb{R}^{W\times H\times C},\; z\in\mathcal{Z},
\)
where $W$ and $H$ denote the width and height of the input image, $C$ the number of channels, and $z$ encodes objects and lifted predicates over their attributes and relations. This non-injective mapping can severely hinder learning, particularly when the policy must distinguish between perceptually different but symbolically identical states.

\medskip
\textbf{Example 1.}\textit{ In Atari game Kangaroo, consider a simplified environment, where there is no enemy (no thrown coconut or monkeys) and the goal is for the mother kangaroo to reach its baby at the top level through some ladders, as shown in Figure~\ref{fig:flowchart} (top-left). One intuitive way to symbolically represent this environment is with three predicates, \texttt{onLadder}, \texttt{leftOfLadder}, and \texttt{rightOfLadder}. In this setting, if the player is at the left of a ladder and \(n\) steps away from the ladder, there are at least \(6^n \) visual configurations that are mapped to the same \texttt{leftOfLadder} predicate, given that the player can face right or left, sat-down or jumped-up in a captured image.}
\medskip

To address these challenges, we incorporate \textit{human knowledge in the form of expert advice} as a central component. The form of our advice is \emph{action preferences}—a set of preferred actions—and, optionally, a \emph{relational abstraction} that specifies where the advice applies. 



    


  
\begin{definition}\label{def:advice}
Let $z \in \mathcal{Z}$ be a symbolic state described by a conjunction of predicates 
$p_1, \dots, p_n$. An \emph{advice} is a pair
\(
\alpha = (\varphi, A^{\mathrm{adv}}), \; \varphi \in \Lambda,\; A^{\mathrm{adv}} \subseteq \mathcal{A},
\). where $\varphi$ is an (optional) abstraction that maps the original predicates $\{p_1, \dots, p_n\}$ 
to a (possibly equivalent) conjunction of lifted predicates, and $A^{\mathrm{adv}}$ is the set of preferred 
actions in the state. The abstraction $\varphi$ specifies the critical relational conditions under which the 
advice is valid, while the set $A^{\mathrm{adv}}$ encodes the human’s recommended action choices. If 
no abstraction is provided, $\varphi$ defaults to the conjunction of all predicates describing $z$. We write 
$z \models \varphi$ when the state $z$ satisfies the abstraction $\varphi$.
\end{definition}

\medskip
\textbf{Example 2.} \textit{Consider the simplified (no threats) Atari game \textit{Seaquest}. 
Suppose the agent encounters a problematic state $z^\star$ described by the following set of predicates:}
\begin{align*}
&\texttt{visibleDiver(obj27, s1), visibleDiver(obj26, s1)}, \\
&\texttt{deeperThanDiver(obj1, obj26, s1), deeperThanDiver(obj1, obj27, s1)}, \\
&\texttt{oxygenLow(obj36, s1), facingLeft(obj1, s1)}, \\
&\texttt{rightOfDiver(obj1, obj27, s1), rightOfDiver(obj1, obj26, s1)}.
\end{align*}

\textit{In this situation, the expert advises the agent to take the action \texttt{move\_up} in order to replenish oxygen. 
Formally, the advice is given as:}
\(
\alpha = (\varphi, A^{\mathrm{adv}}), \;
\varphi = \{\texttt{oxygenLow}(\cdot)\}, \;\; A^{\mathrm{adv}}=\{\texttt{move\_up}\}.
\) \textit{Here, the abstraction $\varphi$ retains only the predicate \texttt{oxygenLow}, ignoring the presence of divers, orientation, or spatial relations. 
Thus, the advice generalizes: whenever the agent observes that oxygen is low—regardless of other predicates in the state—it should prioritize moving up to refill oxygen.}
\medskip

Note that while the example lists a single action in the advice, the framework accepts a set of actions (preferred) and treats the rest of the actions as non-preferred actions according to the advice.

Figure~\ref{fig:flowchart} illustrates the overall workflow of our proposed framework, which applies \textit{Relational Fitted Q-learning (RFQ)} (\cite{das2020fitted}) directly to both structured domains and raw image environments using symbolic representations extracted with object-centric tools such as \texttt{OC-Atari} (\cite{delfosse2024ocatari}). The key idea is to transform pixel-based observations into lifted symbolic representations that capture object attributes and relations. We then train an RFQ model \(
Q:\mathcal{Z}\times\mathcal{A}\to\mathbb{R},\; Q(z,a),
\) on this symbolic representation to induce a relational policy. 

A second central feature of this framework is the process of \emph{active advice elicitation}. 
The agent continuously monitors its own learning progress and detects \emph{problematic states}—regions of the state space where the policy exhibits high uncertainty. We calculate the uncertainty based on the policy entropy, which measures how uncertain or stochastic a policy is across its action choices:
\begin{equation}
H(s) = -\sum_{a \in \mathcal{A}} \pi(a|s) \log \pi(a|s),  
\label{eq:policy_entropy}
\end{equation}
where \( \pi(a|s) \) is the probability of selecting action \( a \) in state \( s \), and \( \mathcal{A} \) denotes the action set.

\begin{algorithm}[t]
\caption{Relational Active Advice Elicitation in Reinforcement Learning (RAEL)}
\label{alg:rael}
\DontPrintSemicolon

\SetKwInOut{Input}{Input}
\SetKwInOut{Output}{Output}

\Input{%
  Object-centric map \(f\);
  exploration probability \(p_{\epsilon}(t)\);\\
  advice taking probability \(\rho(t)\); 
  monotone function \(\psi\);\\
  advice budget \(B\); training rollouts \(N_{\mathrm{train}}\); evaluation rollouts \(N_{\mathrm{eval}}\).%
}
\Output{Learned \(Q\in\mathcal{F}\) and policy \(\pi(z)=\arg\max_a Q(z,a)\).}

Initialize \(Q\in\mathcal{F}\) (to a constant), data  buffer \(\mathcal{B}\leftarrow\emptyset\), advice memory $\mathcal{A}dv \leftarrow \emptyset$\ , Advice Budget $B$;

\For{$\text{iter} = 1,2,\dots$}{
    \For{$e=1$ \KwTo $N_{\text{train}}$}{
        Reset env, observe $x_0$, set $z_0 \leftarrow f(x_0)$\;
        \While{episode not terminal}{
            Sample \(u\sim\mathrm{Uniform}[0,1]\)\;

            \uIf{\(u < p_{\epsilon}(t)\)}{
        \(a_t \sim \mathrm{Uniform}(\mathcal{A})\); 
      }
      \uElseIf{ \(\exists(\varphi,a^{\mathrm{adv}})\in\mathcal{A}\mathrm{dv}: z_t\vDash\varphi\) \ \textbf{and} \ \(u < \rho(t)\) }{
        \(a_t \leftarrow a^{\mathrm{adv}}\); 
      }
      \Else{
        \(a_t \leftarrow \arg\max_{a} Q(z_t,a)\); 
      }
            Execute $a_t$;
            Compute $q_t$ using Equations (\ref{eq:bellman}) and (\ref{eq:proj_after});
            Store $(z_t,a_t,q_t)$\ in $\mathcal{B}$;
        }
    }
    
    \tcp{Relational Fitted Q-update with advice boosting}
    Learned gradient boosted relational regression trees to fit $\mathcal{B}$ and obtain the new $Q$ function;

    \tcp{Compute uncertainty and query expert}
    
  \(\mathcal{E}\leftarrow\text{Rollout}(Q, N_{\mathrm{eval}})\)\;
  
  Compute \(H(z) = -\sum_{a}\pi(a\mid z)\log\pi(a\mid z)\) for all \(z\in\mathcal{E}\)\;
 
  \(z^\star \leftarrow \arg\max_{z\in\mathcal{E}} H(z)\)\;
  
  \If{\(B>0\)}{
    Query expert on \(z^\star\)\ to get \((\varphi, a^{\mathrm{adv}})\)
    
    Insert \((\varphi,a^\mathrm{adv})\) into \(\mathcal{A}\mathrm{dv}\)\;
    
    \(B \leftarrow B - 1\)\;
  }
}
\end{algorithm}

The overall working of our RAEL framework is summarized in Algorithm~\ref{alg:rael}. 
After each iteration, the agent performs an evaluation roll-out and collects visited symbolic states $\mathcal{D}_{\text{eval}}=\{z\}$. It then selects the most uncertain state \( z^\star \in \arg\max_{z\in\mathcal{D}_{\text{eval}}} U(z)\), and given  an advice budget $B\in\mathbb{N}$, it seeks advice for that state from the expert, and stores the advice set in a memory set $\mathcal{A}\mathrm{dv}\subseteq \Lambda\times\mathcal{A}$ (Definition \ref{def:advice}).
During training, action selection at $z_t$ follows the below Probabilistic Policy Reuse(PPR) scheme \citep{FernandezVeloso2006}, to ensure that the expert advice is applied selectively, while preserving exploration and exploitation:

\begin{equation}
a_t \sim
\begin{cases}
\mathrm{Uniform}(\mathcal{A}) & \text{if } u<p_{\epsilon}(t),\\[2pt]
a^\mathrm{adv} & \text{if }\exists\,(\varphi,a^\mathrm{adv})\in\mathcal{A}\mathrm{dv}\ \text{s.t. }z_t\vDash\varphi\ \text{and } u<\rho(t),\\[2pt]
\arg\max_{a\in\mathcal{A}} Q(z_t,a) & \text{otherwise},
\end{cases}
\label{eq:ppr}
\end{equation}
where $u\sim\mathrm{Uniform}[0,1]$, \( \rho(t) \in [0, 1] \) is the advice-taking probability which controls the degree of expert influence, and $p_\epsilon(t)$ is the exploration rate. Both $\rho(t)$ and $p_\epsilon(t)$ can decay with $t$. Using PPR, we ensure that the advice acts as a soft constraint on the learner’s policy rather than a hard instruction. Hence, if advice is noisy or suboptimal, the agent is not forced to follow it and can adapt based on environmental feedback.

When an advised action ($a_{\text{adv}}$) is used in a state $z_t$ that satisfies the advice condition, we first perform the standard Bellman optimality update
\begin{equation}
Q(z_t,a_{\text{adv}}) \leftarrow (1-\alpha)Q(z_t,a_{\text{adv}}) + \alpha y^{\text{adv}}_t\,
\label{eq:bellman}
\end{equation}
where \(y^{\text{adv}}_t = r^{\text{adv}}_t + \gamma \max_{a'} Q(z_{t+1},a') \; \forall a \in \mathcal{A}\), and $r^{\text{adv}}_t$ is the reward received after taking the advised action $a^{\mathrm{adv}}$ at time $t$. We then project the advised action’s value to ensure it dominates other actions at $z_t$ by at least a small margin $\delta>0$:
\begin{equation}
q_t = Q^{\text{adv}}(z_t,a_t) =
\begin{cases}
Q(z_t, a_t) & \text{if} \ a_t  \ne a_{\text{adv}},\\
\max\Big(Q(z_t,a_{\text{adv}}), \max_{a \neq a_{\text{adv}}} Q(z_t,a) + \delta\Big) &  a_t = a_{\text{adv}},
\end{cases}
\label{eq:proj_after}
\end{equation}



\noindent
Thus, in short, Algorithm \ref{alg:rael} alternates between (i) training episodes where the agent explores the environment, executes actions, and stores transitions following the expert guidance (ii) fitted Q-updates where relational regression trees are trained based on the training trajectory and (iii) evaluation rollouts where policy entropy is monitored to identify the most uncertain states. If the advice budget has not been exhausted, the most uncertain state is presented to the expert, who provides a set of preferred actions optionally paired with a lifted abstraction. The advice is stored in the memory set $\mathcal{A}dv$ and influences subsequent action selection according to Eq.\ref{eq:ppr}. The boosted Q-values  (Eq.\ref{eq:proj_after}) ensures that expert-advised actions dominate in the corresponding states.

\section{Experiments and Results}

To evaluate the effectiveness of our proposed method, we designed a set of experiments addressing the following core questions:
\begin{itemize}
    \item \textbf{Q1: Can RAEL achieve sample-efficient learning?} We hypothesize that our method offers improved sample efficiency by directly learning symbolic policies. Also, compared to approaches that use neural guidance, we hypothesize that our method shows better policy fidelity, especially when the base rules are imperfect or misaligned in neural-guided methods.
    \item \textbf{Q2: Can active advice improve learning outcomes?} We aim to assess whether enabling the agent to selectively query for guidance in uncertain states leads to better policies.
    
    \item \textbf{Q3: Is adding abstraction guidance more informative than simple action preferences?} We compare two types of expert interventions: (i) \textit{action preference} advice, where the expert suggests the optimal action(s), and (ii) \textit{abstraction-based} advice, where the expert additionally highlights which predicates of the state influenced their decision.

\end{itemize}
\textbf{Environments.} We evaluated our approach across both subsymbolic and symbolic domains. For the former, we conducted experiments on two Atari games---\textbf{Kangaroo} and \textbf{Seaquest}---using the Atari Learning Environment benchmark \citep{bellemare2013arcade}, which is widely adopted in reinforcement learning research, particularly in relational reasoning tasks. We use OC-Atari \citep{delfosse2024ocatari} to extract the symbolic representation from raw pixels. To facilitate clearer analysis and interpretation, we did the experiments in simplified environments. For this purpose, we have used Hackatari \citep{delfosse2024a} to decrease the threats in the environment. For example, the thrown coconuts in Kangaroo and missiles in Seaquest are eliminated. Although the method was developed and evaluated within a simplified setup, its applicability to more complex environments remains to be explored. Prior research suggests that relying on logical reasoning in high-risk or time-critical situations is often impractical, as such situations require rapid decision-making. In these cases, neural policies are preferred for their fast response, while logical reasoning can be reserved for lower-risk states \citep{blendrl}.
To examine the effectiveness of our active advice mechanism in structured symbolic domains, we additionally evaluated on the \textbf{Blocks World} (Stack task) \citep{slaney2001blocks}. In this environment, blocks are arranged into towers, and the agent’s goal is to construct a single tower by stacking all blocks. We defined a positive reward for reaching the goal and a small negative reward for intermediate steps. In Atari environments, we used a large negative reward for terminal failure states (e.g. being punched by a monkey in Kangaroo or running out of oxygen in Seaquest). In our experiments, we set the advice budget to $3$ for the simpler Kangaroo domain and $5$ for the more complex Seaquest and Blocks World domains. While we treated the budget as a fixed hyperparameter in this work, quantifying this parameter systematically is an interesting direction for future research.

\noindent \textbf{Evaluation Metric.} We  used two metrics for our evaluation-- \textit{cumulative rewards} and \textit{policy fidelity}. Cumulative reward is computed by executing the agent  with its learned policy over multiple evaluation episodes and averaging the total rewards across the trajectories.
Policy fidelity is measured by comparing the learned policy against a reference set of rules provided by an external oracle. In our case, we treat the oracle as an expert source—such as a domain specialist or a large language model—that specifies a set of essential rules believed to be near-optimal for the task. Specifically, we used GPT-5\footnote{https://platform.openai.com/docs/models/gpt-5} to generate these rules, capturing the behaviors it predicts would lead to high performance. After training, the agent’s learned rules are evaluated against this oracle rule set to determine the extent to which the learned policy satisfies the prescribed near-optimal conditions.

\noindent \textbf{Implementation Details.} All Experiments were run on a server with an AMD EPYC 7343 CPU, 256 GB RAM, and an NVIDIA L4 GPU (24 GB). The algorithms were implemented in Python 3.10, with relational components using the RFQ codebase~\citep{das2020fitted} and Atari environments via the OpenAI Gym interface, and OC-Atari object centric representation module.

\subsection*{Q1: Can RAEL achieve sample-efficient learning?}
To evaluate sample efficiency, we compare RAEL against established neuro-symbolic reinforcement learning baselines under limited training steps. Specifically, we consider NUDGE~\citep{nudge23} and BlendRL~\citep{blendrl}.
NUDGE is a symbolic policy learning framework that employs differentiable logic programs as policy functions.
BlendRL extends NUDGE by incorporating neural networks, enabling hybrid policy computations and achieving state-of-the-art performance on the Kangaroo and Seaquest environments as a relational RL baseline.
We additionally compare RAEL against a purely neural baseline: a PPO agent (Neural PPO) that uses a convolutional neural network as its policy function~\citep{schulman2017proximal}. For this, we adopt a publicly available implementation of PPO~\citep{huang2022cleanrl} with their default model architecture and hyperparameters.

Table~\ref{tab:results} presents the mean cumulative rewards after 300k training steps. Overall, RAEL achieved the highest cumulative rewards across both environments. The logic-based baseline, NUDGE, obtained moderate rewards in Seaquest but failed in Kangaroo, highlighting its limitations in handling complex, dynamic scenarios (e.g., nearby enemies) due to the lack of direct structure learning. Baselines with a major neural component, Neural PPO and BlendRL, also failed in both environments, illustrating the sample inefficiency of policies that rely primarily on neural networks, which typically require millions of training steps to converge. In contrast, RAEL exploited sample-efficient structure learning to attain superior performance. Moreover, when augmented with external advice, RAEL improved further, demonstrating not only efficiency but also the capacity to integrate external guidance, which are essential aspects of human intelligence.

\begin{table}[t]
    \centering
        \caption{\textbf{RAEL is sample efficient.} Mean return after 300k training steps. RAEL outperforms purely gradient-based methods, such as NeuralPPO, NUDGE, and BlendRL, which suffer from sample inefficiency. }
    \begin{tabular}{@{}llllll@{}}
    \toprule
    \textbf{Environments} & \textbf{NeuralPPO} & \textbf{NUDGE} & \textbf{BlendRL} & \textbf{RAEL (w/o Abst.)} & \textbf{RAEL} \\ \midrule
    \textbf{Kangaroo}     & $0.0\smallpm{0.0}$  & $0.0\smallpm{0.0}$  & $0.0\smallpm{0.0}$ & 51 & \textbf{55 }\\
    \textbf{Seaquest}     & $0.0\smallpm{0.0}$ & $20.2 \smallpm{18.55}$ & $0.0\smallpm{0.0}$ & 40  & \textbf{110 }\\ \bottomrule
    \end{tabular}
    \label{tab:results}
\end{table}

\subsection*{Q2: Can active advice improve learning outcomes?}
To measure the effectiveness of active advice, we first hypothesized that incorporating expert advice even as a prior rule would aid learning. However, our experiments revealed a counterintuitive outcome: introducing prior advice in a simple, threat-free environment could actually undermine performance. For instance, in the basic \textit{Kangaroo} setup---without extensive reward shaping---the agent was able to learn near-optimal policies on its own. However, when we introduced a seemingly benign piece of advice (e.g., `whenever you are on a ladder, go up''), the learning process deteriorated significantly.
The problem is when we manipulated the Q-values to artificially favor the \texttt{go\_up} action while on the ladder, although expecting it to accelerate convergence, it led to undesirable behavior: the agent began intentionally descending the ladder only to climb back up and repeatedly exploit the inflated Q-value, thereby getting stuck in a loop---a classic pitfall of poorly designed reward shaping. This illustrates that in environments where the dynamics are simple, it is often better to avoid advice altogether than to risk introducing imperfect or misleading guidance.

Table \ref{tab:policy-fidelity} shows that imperfect prior advice can harm performance, even if combined with reward shaping. In the no-threat setting, Reward Shaping (RS), Prior Advice (PA), and no intervention all achieved perfect fidelity (3 out of 3 rules). However, when RS was combined with PA, fidelity dropped to \textbf{2 out of 3}, indicating that even in a simple environment, incorrect or overly general prior rules can conflict with the shaped reward and hinder optimal policy learning. In the environment with the “Monkey” threat, fidelity remained low (1 out of 4) for all setups, suggesting that neither fixed prior advice nor reward shaping alone is sufficient in more challenging conditions. These observations motivate the use of adaptive, on-demand advice, which can provide targeted guidance when needed.

\begin{figure}[ht]
  \centering
  \includegraphics[width=0.65\linewidth]{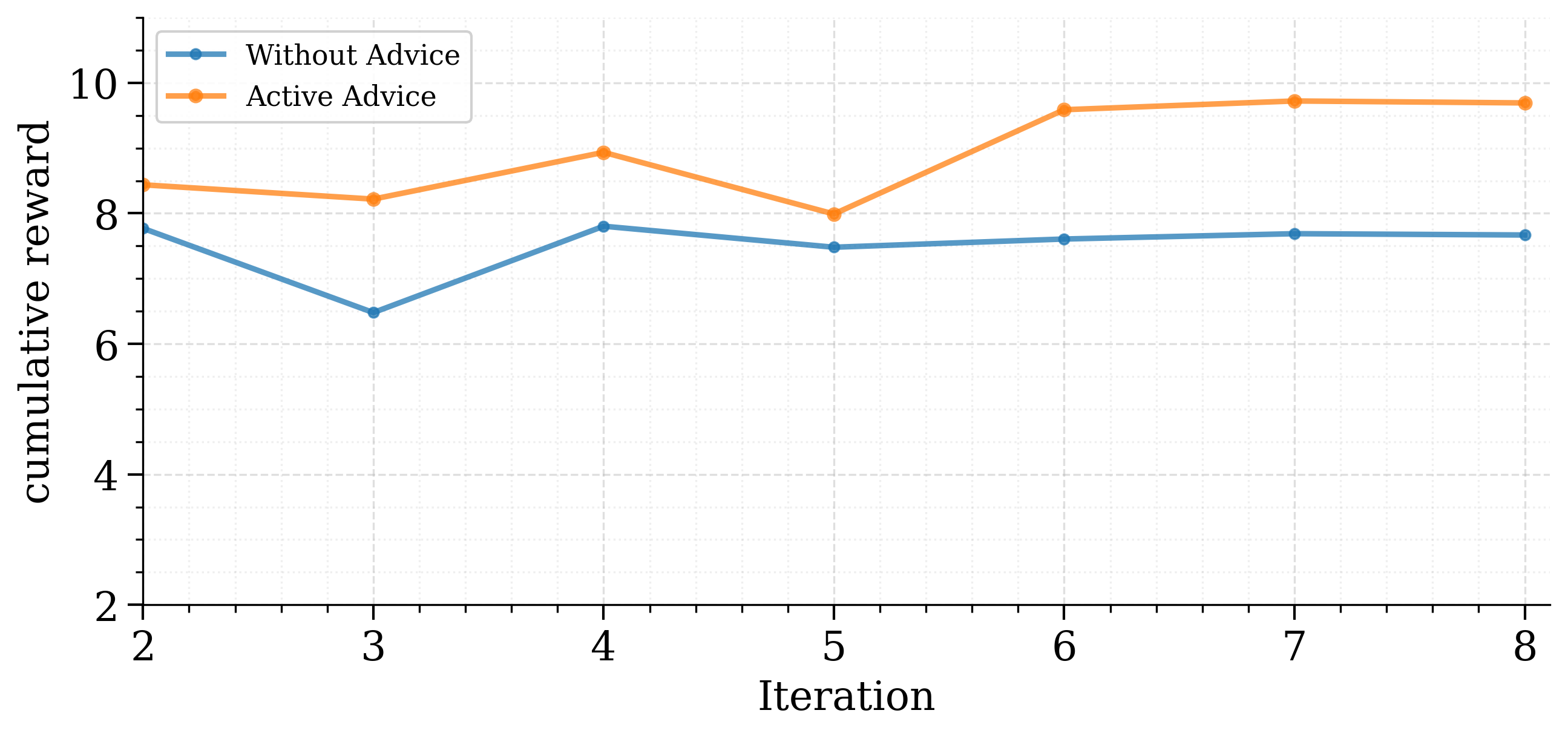}
  \caption{
  \textbf{Active advice improves performance in Blocks World.}
Cumulative reward as a function of training iteration for the stacking task. The orange curve shows mean performance with active advice (entropy-based querying), while the blue curve shows training without advice. 
Active advice consistently yields higher rewards and more stable learning compared to the advice-free baseline.
  }\label{fig:blockResutls}
\end{figure}

Our results on Blocks World environment (Stack task), shows that actively interacting with expert on problematic regions improves learning (Figure \ref{fig:blockResutls}). This proves our hypothesis and shows that our method not only generalizes effectively to unstructured, sub-symbolic environments but also accelerates convergence in structured relational domains.

\begin{table}[t]
\centering
\caption{Policy fidelity for passive (RS, PA, RS+PA, No Advice) and active advice (with/without abstraction) on \textit{Kangaroo} (threat / no threat) and \textit{Seaquest}. “--” denotes not evaluated.
}
\setlength{\tabcolsep}{5pt}
\begin{tabular}{llccc}
\toprule
 & & \multicolumn{2}{c}{\textbf{Kangaroo}} & \textbf{Seaquest} \\
\cmidrule(lr){3-4}\cmidrule(l){5-5}
\textbf{Group} & \textbf{Setting}  & No threat & with Monkey & Simplified \\
\midrule
\multirow{4}{*}{Passive}
  & RS (Reward Shaping)          & 100\% & 25\%  & -- \\
  & PA (Prior Advice)            & 100\% & 25\%  & -- \\
  & RS+PA                        & 67\%  & 25\%  & -- \\
  & No Advice                    & 100\% & 25\%  & 21\% \\
\midrule
\multirow{2}{*}{Active}
  & Active w/o abstraction      & --    & 75\%  & 44\% \\
  & Active w abstraction        & --    & \textbf{100}\% & \textbf{75}\% \\
\bottomrule
\end{tabular}
\label{tab:policy-fidelity}
\end{table}
\subsection*{Q3: Is adding abstraction guidance more informative than simple action preferences?}

To measure the effectiveness of abstraction advice, we evaluated three configurations.

• \textbf{No Advice (Baseline)} – The agent learns without any expert input.
• \textbf{Active Advice w/o Abstraction} – The expert provides only the preferred action(s) when requested.

• \textbf{Active Advice w/ Abstraction} – The expert provides both the preferred action(s) and the relevant state predicates that justify it.

In the abstraction-guided setup, the agent requests advice in states of high uncertainty. Upon receiving the request, the expert evaluates the current state in terms of symbolic predicates and identifies which predicates directly support the advised action. 

\begin{figure}[ht]
  \centering
  \includegraphics[width=\linewidth]{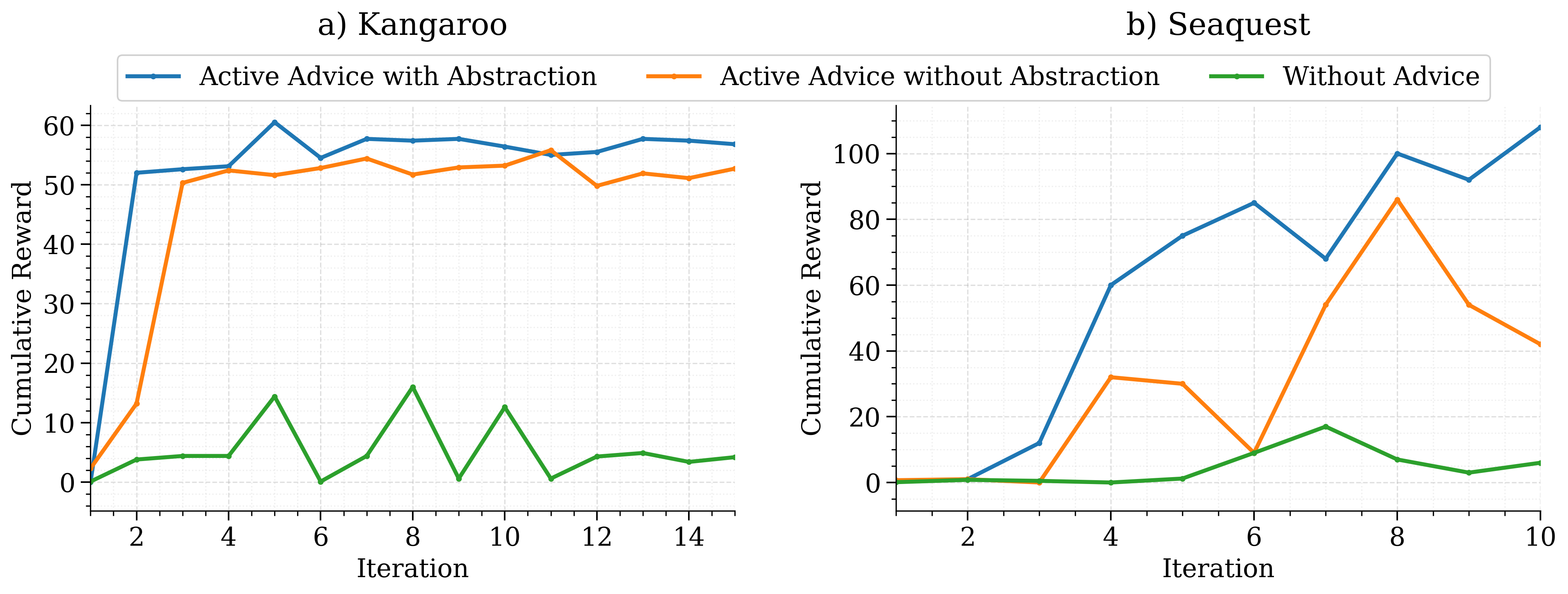}
  \caption{\textbf{Impact of abstraction-guided advice in Atari.}
Cumulative reward over training iterations for (a) Kangaroo and (b) Seaquest. The blue curve shows RAEL with active advice including relational abstraction, the orange curve is active advice with only action preferences, and the green curve is no advice. Across both domains, advice with abstraction yields faster and higher returns, demonstrating improved sample efficiency and policy quality.}\label{fig:atariResutls}
\end{figure}

Figure~\ref{fig:atariResutls} shows the average cumulative return in \textit{Kangaroo} and \textit{Seaquest}. Both active advice models outperform the baseline, confirming that expert guidance accelerates learning. Moreover, incorporating abstractions into advice leads to faster convergence compared to action-only advice, as the agent can generalize the advice across states more effectively when key predicates are specified. For example, in the \textit{Seaquest} environment, when the predicate \textit{oxygenLow} is true, the expert advises the agent to move up to refill oxygen. By abstracting over this predicate, the agent learns that the action remains a priority regardless of other objects present in the state, which reduces the number of queries needed for similar situations.

In addition to performance, we evaluated the policy fidelity of the abstraction-guided model. As shown in Table~\ref{tab:policy-fidelity}, the abstraction-based advice consistently achieves the highest fidelity, followed by action-only advice and then the no-advice baseline. This supports our hypothesis that explicitly highlighting relevant predicates allows the agent to better internalize and apply expert knowledge. 

\section{Related Work}
\paragraph{Relational and Neuro-Symbolic RL.} RRL~\citep{Dzeroski01RelationalRL, kersting2004bellman, kersting2008non, Lang12RelationalRLModel, DeepRelationalRLNeSy} aims to scale RL to environments best described by objects and relations using logical or lifted representations and probabilistic reasoning.
Recent neuro-symbolic methods seek to combine the interpretability of symbolic policies with the expressiveness of deep neural networks, enabling structure aware policies to be learned from raw inputs or partially observed states \citep{JiangL19NLRL, nudge23, DeepRelationalRLNeSy, nesy_markov, blendrl}.
However, most existing approaches rely heavily on gradient-based optimization for policy learning, which makes them sample-inefficient—requiring a large number of training steps to obtain satisfactory policies.
More importantly, these methods are not explicitly designed to align with human preferences or incorporate human guidance.
In contrast, RFQ learning \citep{das2020fitted}, which we use in this work, addresses the above limitations by performing direct structure learning and reducing sample complexity while enabling human-aligned policy induction.

\paragraph{Human Allied RL and Advice Integration.} 
A longstanding goal in RL is to leverage human expertise to improve sample efficiency, safety, and alignment with user preferences, particularly in environments with sparse rewards or large state spaces \citep{Retzlaff2024HumanintheLoop}. {Early work on program-guided Q-learning and advice-taking RL \citep{Andre2001,maclin1996} showed that incorporating high-level, imperative guidance—whether as programmed macro-actions \citep{Andre2001} or as human-provided rules injected into a Q-learner \citep{maclin1996}—can accelerate learning and improve policy quality, providing temporal (sequential) and contextual structure beyond single state–action labels. Building on this, several principled approaches have emerged for incorporating advice via reward shaping \citep{Wiewiora2003}, logic constraints \citep{Baert2023} as well as uncertain opinions \citep{Dagenais2024}. In relational domains, advice can be especially impactful, as shown by \cite{driessens2004integrating}, who incorporated advice into RRL to make Q Learning feasible in structured environments with sparse rewards. 
\cite{Korupolu2012} introduced a relational approach where agents interpret human instructions as either action preferences or state preferences
. \citet{Bignold2021} proposed a persistent rule-based interactive RL method using Ripple-Down Rules and probabilistic policy reuse \citep{FernandezVeloso2006}, retaining advice across episodes to improve efficiency. 
More recently, \cite{Guo2023} proposed an \emph{Explainable Action Advising} framework, where the expert provides both the preferred action as well as an explanation that acts as an abstraction, enabling the agent to reason.
{Compared to these approaches, RAEL focuses on action-preference advice at the state level, optionally paired with symbolic abstractions that highlight relevant predicates. This choice makes the integration simple and tractable within relational fitted Q-learning, while still enabling generalization across states.

The mentioned previous methods rely on \emph{predefined static rules} to improve the sample efficiency. 
Actively seeking advice has previously been explored in the context of inverse RL \citep{odom2015active}. However its utility in direct policy/value learning has not been studied.
Our framework, in contrast to the above works, introduces an active, abstraction-aware advice elicitation framework that selectively queries experts in high-uncertainty states and integrates their guidance directly into relational policy learning process. 

\section{Conclusion}

We considered the challenging problem of learning symbolic policies from structured and unstructured domains in the presence of human expert. The key idea in our work is inspired by active learning that assumes the presence of a human expert who can be queried with a predefined budget. Instead of obtaining specific action labels before learning commences, we solicit for soft constraints on the policy by explicitly computing the uncertainty over the policy, and update the learned policies accordingly. Our algorithm handles unstructured data by constructing a symbolic representation by employing object-centric identification from images. We performed experiments in both structured and unstructured domains and demonstrated the value of employing a rich representation as well as the importance of access to the human expert while learning.

There are several possible directions for future research. Handling both structured and unstructured information in a single domain (i.e., truly multimodal) is an interesting direction. Employing richer forms and modalities of domain knowledge and developing methods that can treat these constraints in an unified manner would allow richer models to be learned. Scaling these methods to larger tasks by developing fully differentiable models can demonstrate the value of neurosymbolic learning. Extending the algorithm to other RL methods to develop a suite of RRL library is an exciting next step. Studying the quantity of advice (e.g., how frequently the agent should interact with a human expert) and the quality of advice (e.g., mechanisms for modeling adviser reliability, adaptively weighting input, and integrating external knowledge), represent complementary avenues for future work. Finally, learning from different domain knowledge sources (both humans and LLMs) in a seamless manner is an exciting direction for future research.

\section*{Acknowledgments}
\emph{SN}, \emph{FG} and \emph{SS} gratefully acknowledge the generous support by the ARO award W911NF2010224 and the DARPA Assured Neuro Symbolic Learning and Reasoning (ANSR) award HR001122S0039.
Additionally, this work has benefited from the early stages of funding by the Deutsche Forschungsgemeinschaft (DFG, German Research Foundation) under Germany’s Excellence Strategy— "Reasonable AI" (EXC-3057) and "The Adaptive Mind" (EXC-3066); funding will begin in 2026.

{\parindent -10pt\leftskip 10pt\noindent
\bibliographystyle{cogsysapa}
\bibliography{format}
}

\end{document}